%% file: ms.tex
\documentclass[10pt]{article}
\input{hicss-packages.tex}
\usepackage{comment}
\usepackage{amsmath,amssymb,amsfonts}
\usepackage{algorithm, algorithmic}
\usepackage{graphicx,epstopdf}
\usepackage{textcomp}
\usepackage{xcolor}
\usepackage{booktabs} 
\usepackage{hyperref}
\usepackage{multirow}
\usepackage{subfig}

\title{Detecting Concept Drift With Neural Network Model Uncertainty}

\author{Lucas Baier \\
IBM, Germany\\
{\underline{ lucas.baier@ibm.com}} \And
Tim Schlör \\
KIT, Germany \\
{\underline{ tim.schloer@gmail.com}} \And
Jakob Schöffer \\
KIT, Germany \\
{\underline{ jakob.schoeffer@kit.edu}} \And
Niklas Kühl\\
KIT, Germany \\
{\underline{ niklas.kuehl@kit.edu}}
}

\date{}

\begin{document}
\maketitle
\begin{abstract}
\noindent
Deployed machine learning models are confronted with the problem of changing data over time, a phenomenon also called \emph{concept drift}.
While existing approaches of concept drift detection already show convincing results, they require true labels as a prerequisite for successful drift detection.
Especially in many real-world application scenarios---like the ones covered in this work---true labels are scarce, and their acquisition is expensive.
Therefore, we introduce a new algorithm for drift detection, \emph{Uncertainty Drift Detection (UDD)}, which is able to detect drifts without access to true labels.
Our approach is based on the uncertainty estimates provided by a deep neural network in combination with Monte Carlo Dropout.
Structural changes over time are detected by applying the \emph{ADWIN} technique on the uncertainty estimates, and detected drifts trigger a retraining of the prediction model.
In contrast to input data-based drift detection, our approach considers the effects of the current input data on the properties of the prediction model rather than detecting change on the input data only (which can lead to unnecessary retrainings).
We show that \emph{UDD} outperforms other state-of-the-art strategies on two synthetic as well as ten real-world data sets for both regression and classification tasks.
\end{abstract}

\subsubsection*{Keywords:}

Concept Drift Detection, Uncertainty, Monte Carlo Dropout, No Labels, Data Stream

\section{Introduction}
Across most industries, machine learning (ML) models are deployed to capture the benefits of the ever-increasing amounts of available data.
When deploying models, most practitioners assume that future incoming data streams are stationary, i.e., the data generating process does not change over time.
However, this assumption does not hold true for the majority of real-world applications (\cite{aggarwal2003framework}).
%, where usually changes in the distribution of the incoming data occur \cite{aggarwal2003framework}.
%
In the literature, this phenomenon is referred to as \emph{concept drift} or \emph{dataset shift}, which usually leads to a decreasing prediction performance (\cite{Gama2014, baier2019cope}).
Even small changes or perturbations in the distribution can cause large errors---which has been shown through, e.g., adversarial examples (\cite{szegedy2014intriguing}).

The concept drift community has developed several learning algorithms that are able to adapt incrementally (\cite{shalev2011online})
%\cite{hulten2001mining, carvalho2006single}
or detect concept drift and trigger retrainings of a corresponding learning algorithm (\cite{bifet2007learning,gama2004learning}).
These techniques usually require full and immediate access to ground-truth labels, which is an unrealistic assumption in most real-world use cases.
As an example, let us consider a manufacturing line with a manual end-of-line quality control.
By collecting sensor data from all manufacturing stations and combining this information with previously acquired quality assessments (labels) of human experts, a predictive model can be built to replace the manual quality control and thus reduce repetitive and expensive human labour.
However, this prediction model is likely exposed to concept drift due to, e.g., modifications in raw materials, machine wear, ageing sensors or changing indoor temperatures due to seasonal changes.
A continuous stream of true labels for concept drift detection is not available in this use case---which is why traditional concept drift detection algorithms are not applicable.

To address these shortcomings, we investigate the following research question: \textit{How can we improve concept drift detection in situations with limited availability of true labels?} To that end, we propose a novel concept drift detection algorithm which detects drifts based on the prediction uncertainty of a neural network at inference time, and we call this method \emph{Uncertainty Drift Detection (UDD)}.
Specifically, we derive uncertainty by applying Monte Carlo Dropout (\cite{gal2016dropout}).
In case of a detected drift, we assume that true labels are available upon request (e.g., provided by domain experts) for retraining of the prediction model.
In contrast to most drift detection algorithms, \emph{UDD} can be used for both regression and classification problems.
We evaluate \emph{UDD} on two synthetic as well as ten real-world benchmark data sets and show that it outperforms other state-of-the-art drift detection algorithms.

\section{Background and Related Work}

\subsection{Dataset Shift and Concept Drift}
\label{sec:cd_definition}
The ML and data mining communities use different terms to describe the phenomenon of changing data distributions over time and its impact on ML models (\cite{moreno2012unifying}).
\emph{Dataset shift} (\cite{quionero2009dataset}) is described as a change in the common probability distribution of input data $x$ and corresponding labels $y$ between training $(tr)$ and test time $(tst)$: $P_{tr}(x,y) \neq P_{tst}(x,y)$.
This is similar to a common definition of \emph{concept drift} (\cite{Gama2014}): $P_{t_0}(x,y) \neq P_{t_1}(x,y)$, where $t_0$ and $t_1$ are two different points in time with $t_1>t_0$.
Note the difference regarding the indices: Dataset shift focuses on the difference between training and testing environment, whereas concept drift refers to the temporal structure of the data.
%and is therefore closely linked to the problem of ML in a data stream setting.

Dataset shift and concept drift can be further divided into different subcategories: \emph{Virtual drift} (\cite{Gama2014}) refers to changes in the distribution of the input data $x$, without affecting the distribution of labels: $P_{tr}(x) \neq P_{tst}(x)$ and $P_{tr}(y \vert x) = P_{tst}(y \vert x)$.
%
%Within the concept drift community, this is also called \emph{virtual drift} \cite{Gama2014}.
%
\emph{Real concept drift} refers to any changes in $P(y \vert x)$, independent of whether this change is triggered by $P(x)$ or not.

\subsection{Handling Concept Drift}\label{sec:handlingdrift}
%While the above terms all describe related topics, research in the area of concept drift not only deals with the detection of distributional changes and their impact on the prediction quality but also focuses on the question of how to adapt and retrain the affected model (\cite{Gama2014}).

There are many reasons for changing data.
%in the context of a ML model.
%
Usually, it is intractable to measure all confounding factors---which is why those factors cannot directly be included in the ML model.
Often, those factors are considered as ``hidden context'' of the ML models' environment (\cite{tsymbal2004problem}).
Concept drift handling has been applied in a variety of different application domains such as spam detection (\cite{Gama2014}) or demand prediction (\cite{baier2021handling}).
%However, there is another line of research that investigates when a concept drift is caused by the predictions of the corresponding ML model itself \cite{baier2019cope,perdomo2020performative}, i.e., the prediction leads to a change in the data distribution.
%
%This can be illustrated with a predictive policing example: An algorithm will predict neighborhoods with a high likelihood of crimes, and police will increase their presence accordingly.
%
%At the same time, more police presence in those neighborhoods may lead to less crime, thereby invalidating the original prediction of the algorithm.
%
In general, three different categories for detecting concept drift can be distinguished (\cite{lu2018learning}): 
First, error rate-based drift detection, which is also the largest group of methods (\cite{lu2018learning}) and aims at tracking changes in the error rate of a ML model.
Popular algorithms in this category are the \emph{Drift Detection Method} (DDM) (\cite{gama2004learning}), \emph{Page-Hinkley} test (\cite{page1954continuous}), and \emph{ADaptive WINdowing} (ADWIN) (\cite{bifet2007learning}).
Note that the error rate-based drift detection necessarily requires access to ground-truth labels.
Second, data distribution-based drift detection usually applies some distance function to quantify the similarity between the distributions of a reference batch of data and the current data.
Algorithms in this category work on the input data $x$ only and do not require true labels for drift detection.
Popular approaches are based on tests for distribution similarity, such as Kolomogorov-Smirnov test (\cite{raab2020reactive}).
%
%Dimensional reduction techniques, such as PCA \cite{rabanser2019failing}, have also been applied.
%
Third, the multiple hypothesis test category detects drift by combining several methods from the previous two categories. 

%Closely related to the evaluation of many concept drift scenarios is the term \emph{online learning}.
%
%Online learning \cite{shalev2011online} in general assumes the following setting: Receive $x_t$, predict $\widehat{y}_t$, receive $y_t$, and compute a loss.
%
%This way, the algorithm can learn from its previous experience and update the parameters accordingly.
%
%This prequential setup (test, then train) is also prevailing in the evaluation of most concept drift methods \cite{vzliobaite2014controlled}. 
%
In many real-world applications, the assumption that all true labels are available is unrealistic (\cite{krawczyk2017ensemble}).
Furthermore, the acquisition of true labels from experts (e.g., in quality control) is likely expensive.
Those limitations have inspired research on handling concept drift under limited label availability.
In general, methods can be distinguished based on their (non-)requirement of true labels for either drift detection or for retraining of the corresponding model:
The first category of algorithms assumes that true labels are available for both drift detection and retraining, but they are only provided in limited portions at specific points in time.
In this category, algorithms based on active learning have been developed, where true labels for selected samples are acquired based on a certain decision criterion (\cite{fan2004active,zliobaite2013active}).
%
%Other approaches under limited label availability apply semi-supervised learning methods with clustering techniques to derive concept clusters which can be investigated for drifts (\cite{masud2012facing,wu2012learning}).
%
The second category requires no true labels for detection of concept drifts, but it uses them for retraining of the model in case of a drift.
One approach uses confidence scores produced by support vector machines during prediction time and compares those over time (\cite{lindstrom2013drift}).
If there is a large enough difference, the model is retrained using a limited set of current true labels.
Other algorithms monitor the ratio of samples within the decision margin of a support vector machine for change detection (\cite{sethi2015don}).
An incremental version of the Kolmogorov-Smirnov test has also been applied in this category (\cite{dos2016fast}).
The third category handles concept drift without any label access, neither for drift detection nor for retraining, e.g., by applying ongoing self-supervised learning to the underlying classifier (\cite{sun2020test}).

Note that the first category requires some true labels continuously over time in order to be able to detect a drift and trigger corresponding retraining.
In contrast, the second category monitors the data stream for drifts based on the input data only and then requires true labels in case a drift has been detected. This is also the category that \emph{UDD} belongs to.
The third category can adapt without any true label knowledge.
However, this category of algorithms also has the least adaption capabilities due to its limited knowledge of changes.

%\cite{souza2020challenges} discusses the importance for real-world evaluation. Problems with Electricity, Forest Covertype and Pokerhand. \cite{zliobaite2013good} confirms this problem, Gas sensor also difficult

\subsection{Uncertainty in Neural Networks}
In many applications it is desirable to understand the certainty of a model's prediction.
Often times, class probabilites (e.g., outputs of a softmax layer) are erroneously interpreted as a model's confidence.
In fact, a model can be uncertain in its predictions even with a high softmax output for a particular class (\cite{gal2016uncertainty}).
Generally, neural networks are not good at extrapolating to unseen data (\cite{haley1992extrapolation}).
Hence, if some unusual data is introduced to the model, the output of a softmax layer can be misleading---e.g., unjustifiably high.
This likely happens in the case of concept drifts.

Generally, existing literature distinguishes two types of uncertainty: \emph{aleatory} and \emph{epistemic} (\cite{der2009aleatory}).
The former (also called \emph{data uncertainty}) can usually be explained by randomness in the data generation process and, e.g., corresponds to the error term in a regression setting.
The latter (\emph{statistical} or \emph{model uncertainty}) usually results from insufficient training data. For classification tasks, uncertainty can be for instance quantified through entropy, variation ratios or mutual information (\cite{hemmer2020deal}).

One state-of-the-art approach to capture model uncertainty for neural networks is \emph{Monte Carlo Dropout} (MCD) (\cite{gal2016dropout}).
While dropout at training time has been widely used as a regularization technique to avoid overfitting (\cite{srivastava2014dropout}), the idea of MCD is to introduce randomness in the predictions using dropout at inference time.
This allows to deduce uncertainty estimates by performing multiple forward passes of a given data instance through the network and analyzing the resulting empirical distribution over the outputs or parameters.

Another family of methods to quantify predictive uncertainty is called \emph{Deep Ensembles} (\cite{lakshminarayanan2017simple}).
In essence, the authors of this paper propose to enhance the final layer of a neural network such that the model's output is not just a single prediction but a set of distributional parameters, e.g., the mean %$\hat{\mu}$
and variance
%$\hat{\sigma}^2$
for a Gaussian distribution.
The corresponding parameters can then be fitted by using the (negative) log-likelihood as loss function.
For previously unseen data, the approach suggests then to train an ensemble of several neural networks with different initializations at random.
The average of all variance estimates can eventually be interpreted as model uncertainty.

Other recent approaches for quantifying uncertainty in neural networks include variational inference (\cite{blundell2015weight}), expectation propagation (\cite{hernandez2015probabilistic}), evidential deep learning (\cite{sensoy2018evidential}), some of which have been applied to areas like active learning (\cite{hemmer2020deal}) and others.
A good overview of state-of-the-art methods for quantifying uncertainty, including an empirical comparison regarding their performance under dataset shift, is provided by Ovadia et al. (\cite{ovadia2019can}).%---however, they do not consider the problem of model retraining.

%\url{https://www.inovex.de/blog/uncertainty-quantification-deep-learning/}

\section{Methodology} \label{section:methodology}
%\begin{itemize}
%\item Explain setup with few labels
%\item Explain uncertainty approach, entropy formula, refer to different approaches for measuring %uncertainty
%\item Explain how ADWIN used on uncertainty
%\item Explain retraining strategies.
%\item Use confidence?
%\item Pseudocode for approach
%\end{itemize}

When labels are expensive and their availability is limited, popular drift detection algorithms like ADWIN, DDM and Page-Hinkley are not applicable in their original form, as these algorithms detect drifts based on a change in the prediction error rate (and therefore require true labels).
As described in Section \ref{sec:handlingdrift}, there are different scenarios for concept drift handling with limited label availability.
In this paper, we develop a novel approach which detects drifts without access to true labels---yet it requires labels for retraining the model.
For detecting drifts, we rely on the uncertainty of a (deep) neural network's predictions.
Previously, it has been shown that the uncertainty of a prediction model is correlated with the test error (\cite{kendall2017uncertainties,roy2018inherent}).
Thus, we argue that model uncertainty can be used as a proxy for the error rate and should therefore be a meaningful indicator of concept drift.
%, since most concept drift detectors are based on the error rate in a stream setting.   

To investigate this hypothesis, we develop the following approach: For each data instance, we measure the uncertainty of the corresponding prediction issued by the neural network.
Subsequently, this uncertainty value is used as input for the ADWIN change detection algorithm.
We call our approach \emph{Uncertainty Drift Detection (UDD)}.
By applying \emph{UDD}, we can detect significant changes in the mean uncertainty values over time.
If a drift is detected, we require true labels for retraining of the model.
Since there are methods for measuring uncertainty in both regression and classification settings, this approach allows to detect concept drifts for both learning tasks---as opposed to most other concept drift detection algorithms, which handle classification tasks only (\cite{krawczyk2017ensemble}).
Note that \emph{UDD} cannot detect any label shift where $P_{tr}(x) = P_{tst}(x)$ and $P_{tr}(y \vert x) \neq P_{tst}(y \vert x)$.
However, we assume that in most real-world settings there is no label shift without any changes in the input distribution.
%
%This is why we evaluate our method on well-known real-world concept drift datasets.
%
%Naturally, simulated datasets with label shift only do not represent an adequate test set.

For drift detection without true label availability, input data-based drift detection, such as Kolmogorov-Smirnov (\cite{raab2020reactive}), is generally also appropriate.
However, considering solely input data bears the risk of detecting changes in features that may not be important for the prediction model.
Specifically, it may occur that input data-based methods detect drifts where no retraining is required, because this drift will have little or no impact on the predictions of the model (e.g., virtual drift where $P_{t_0}(x) \neq P_{t_1}(x)$ and $P_{t_0}(y \vert x) = P_{t_1}(y \vert x)$).
%
%\emph{UDD}, on the other hand, detects only changes in the input data that also have an impact (as reflected by the uncertainty) on the predictions.
%
However, by using uncertainty of the underlying prediction model, we are investigating the effect of input data on properties of the prediction model rather than considering the input data only.
Thus, only changes in the input data distribution relevant to the prediction model are detected.
Imagine a feature in a high-dimensional feature space which is irrelevant for a neural network at inference time (e.g., low or zero weights have been assigned to this feature during training).
An input data-based method will detect a significant change in this feature, even though this drift will not influence the prediction of the model due to the properties of the corresponding weights.
In fact, a detected drift will lead to the acquisition of new labels at a high cost, even though no retraining is required at this point in time.
\emph{UDD}, in contrast, considers only changes in the input data that also have an impact (reflected by the uncertainty) on the prediction model.

For measuring uncertainty and computing predictions, we apply Monte Carlo Dropout (MCD) because it showed the best performance during our experiments. Furthermore, MCD has been shown to work well in a variety of different machine learning tasks (\cite{gal2016uncertainty}) and the computational requirements are limited, which is an important factor in a stream setting. 
%
%During our experiments, MCD showed the best performance. 
%
However, note that the proposed method can be easily extended to use other uncertainty estimates (e.g., Deep Ensembles) as well.
In practice, MCD applies dropout at inference time with a different filter for each stochastic forward pass through the network.
We denote $T$ the number of stochastic forward passes.
%
%and set $T=50$ for regression tasks and $T=25$ for classification tasks.
%
%For regression tasks, we apply a deep feed forward network with three hidden layers $(128, 64, 32)$.
%
%For classification tasks, we vary the structure between three to five hidden layers $[(128, 64, 32, 16, 8), (64, 32, 16, 8), (32, 16, 8)]$.
%
%Each hidden layer is followed by a dropout layer with dropout rate $0.1$.
%
Predictions $\widehat{p}(y \vert x)$ are computed by averaging the predictions for each forward pass $T$ given the samples $w_i$ of model parameters from the dropout distribution and the input data $x$:
\begin{equation}
\widehat{p}(y \vert x) = \frac{1}{T} \sum_{i=1}^{T} p_i(y \vert w_i, x)\,.
\end{equation}
Regression and classification require different methods for determining predictive uncertainty.
We choose to evaluate the uncertainty for \emph{classification} tasks based on Shannon's entropy $H$ over all different label classes $K$:
\begin{equation} \label{e:entropy}
H\left[\widehat{p}(y \vert x)\right] = - \sum_{k=1}^{K} \widehat{p}(y = k \vert x) * \log_{2}{\widehat{p}(y = k \vert x)}\,.
\end{equation}
For \emph{regression} tasks, uncertainty estimates can be obtained by computing the variance of the empirical distribution of the $T$ stochastic forward passes through the network (\cite{gal2016dropout}):
\begin{equation}
    \widehat{\sigma}^{2} = \frac{1}{T} \sum_{i=1}^{T} \left(p_i(y \vert w_i, x) - \widehat{p}(y \vert x)\right)^2\,.
\end{equation}

%At first, data instances with corresponding ground-truth labels have to be gathered for an initial model training.
%
%Here, we use the first five percent of a data stream's instances.
%
%To detect changes in the uncertainty, we utilize the ADWIN change detector.
%
For change detection, we choose ADWIN as it as able to work with any kind of real-valued input and does not require any knowledge regarding the input distribution (\cite{bifet2007learning}).
Other drift detection algorithms such as DDM (\cite{gama2004learning}) or EDDM (\cite{baena2006early}) are designed for inputs with a Binomial distribution and are therefore not applicable to uncertainty measurements (which can have different distributions by nature).
Real-world data streams for concept drift handling are heterogeneous, e.g., in their number of class labels and size (\cite{souza2020challenges}).
This variability is also reflected by heterogeneous distributions of the respective uncertainty indicator.
Furthermore, due to different approaches for computing uncertainty, this indicator varies significantly in scale and fluctuation between regression and classification problems.
Therefore, ADWIN has to be adjusted to each data stream, which can be achieved by setting its sensitivity parameter $\alpha \in (0,1)$:
A change is detected when two sub-windows of a recent window of observations exhibit an absolute difference in means larger than $\alpha$.

% \begin{algorithm}
% \caption{ADWIN Parameter Tuning}
% \label{alg:alpha}
% \begin{algorithmic}[1]
% \STATE{\textbf{Input}: Trained Prediction Model M, DataStream}
% \STATE{\textbf{Output}: ADWIN Change Detector}
% \STATE{\textbf{Initialize}: ADWIN Change Detector, ADWIN $\alpha=0.1$, TuningData}

% \REPEAT
% \STATE{Receive incoming instance $X_t$}
% \STATE{TuningData $\gets$ TuningData $\ \cup \ \{X_t\}$}
% \UNTIL{size(TuningData) $>$ requiredTuningData}
% \STATE{$\hat{y}_{TD}, uncertainty_{TD} \gets$ M.predict(TuningData)}
% \REPEAT
% \FOR{i in $uncertainty_{TD}$}
% \STATE{Add $uncertainty_i$ to ADWIN}
% \IF{ADWIN detects change}
% \STATE{drifts $\gets$ drifts $\cup \ \{i\}$ }
% \ENDIF
% \ENDFOR

% \IF{size(drifts) $=0$ and $\alpha=0.1$}
% \STATE $\alpha=0.002$
% \STATE \textbf{break}
% \ELSIF{size(drifts) $>1$}
% \STATE Decrease ADWIN $\alpha$
% \ELSIF{size(drifts) $<1$}
% \STATE Increase ADWIN $\alpha$
% \ENDIF
% \UNTIL{size(drifts)$=1$}

% \end{algorithmic}
% \end{algorithm}

New data instances arrive individually and are predicted at the time of arrival.
The obtained uncertainty $U_t$ (either expressed as entropy or variance) from the prediction at time $t$ is used as input for an ADWIN change detector.
Once a drift is detected, a retraining of the prediction model is performed.
For retraining, \emph{UDD} uses the most recent data instances in addition to the original training data.
This way, we can ensure that the model (a) can adapt to new concepts and (b) has enough training data for good generalization.
Algorithm \ref{alg:streameval} on page \pageref{alg:streameval} describes the required steps for \emph{UDD} in a regression ($U_t$ equals variance of prediction $\widehat{\sigma}^{2}$) or classification setting ($U_t$ equals entropy of prediction $H_t$).
%
%Note that the proposed method can be applied to any prediction algorithm that is capable of computing an uncertainty estimation.

\begin{algorithm}[t]
\caption{Uncertainty Drift Detection}
\label{alg:streameval}
\begin{algorithmic}[1]
\STATE{\textbf{Input}: Trained model $M$; Data stream $\mathcal{D}$; Training data $\mathcal{D}_{tr}$}
\STATE{\textbf{Output}: Prediction $\widehat{y}_t$ at time $t$}

\REPEAT
\STATE{Receive incoming instance $x_t$}
\STATE{$\widehat{y}_t, U_t \gets$ $M$.predict($x_t$)}
\STATE{Add $U_t$ to $ADWIN$}
\IF{$ADWIN$ detects change}
\STATE{Acquire most recent labels $y_{recent}$}
\STATE{$M$.train($\mathcal{D}_{tr}$ $\cup$ $\mathcal{D}_{recent}$)}
\ENDIF
\UNTIL{$\mathcal{D}$ ends}
\end{algorithmic}
\end{algorithm}

\section{Experiments}
\label{sec:experiments}
For evaluation purposes, we conduct extensive experiments to compare \emph{UDD} with several competitive benchmark strategies on two synthetic and ten real-world data sets.
This stands in contrast to most concept drift literature, where new methods are mainly evaluated on simulated data sets with artificially induced concept drifts.
The code for our experiments can be found under \url{https://github.com/anonymous-account-research/uncertainty-drift-detection}.

\subsection{Experimental Setup}
\label{sec:experimental_setup}
Throughout the experiments, for MCD, we set the number of stochastic forward passes $T=100$ for regression tasks and $T=50$ for classification tasks.
Regarding the deep feed forward network, we vary the structure between three to five hidden layers with relu activation functions depending on the data set.
Each hidden layer is followed by a dropout layer with dropout rate $0.1$ or $0.2$, as it is proposed in the original MCD paper (\cite{gal2016dropout}).
%
%Details regarding the experimental setup for each data set can be found in Table \ref{table:experiment_details} in the appendix.

\begin{figure}[b]
\centering
\includegraphics[width = 1\linewidth]{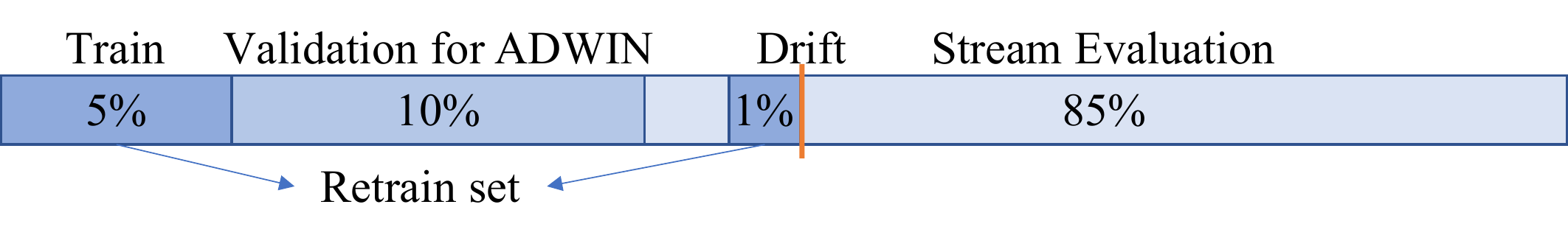}
\caption{Partitioning of data stream.}
\label{fig:stream}
\end{figure}

For initial model training, we use the first five percent of a data stream's instances.
We perform a parameter optimization for \emph{UDD} by requiring the associated ADWIN algorithm to detect one drift on a given validation data set---this yields a concrete value for the sensitivity parameter $\alpha$.
If no drifts are detected on the validation data with the initial value for $\alpha$, we assume that no drifts are present in the validation data and $\alpha$ is set to the \texttt{scikit-multiflow} (\cite{skmultiflow}) default value of 0.002.
We use the ten percent of instances following the initial training data as validation data.
Every time we detect a drift, we provide the last data instances as well as corresponding labels equivalent to one percent of the overall data stream's length.
The exemplary partitioning of a data stream is depicted in Figure \ref{fig:stream}.

In order to benchmark \emph{UDD}, we compare it against six different strategies within two groups. The first group of strategies handles concept drift with \emph{Limited Label Availability} whereas the second group of strategies allows for \emph{Unlimited Label Availability}.

\subsubsection{Limited Label Availability}
%We start by explaining the four strategies within the first group.
%
%The first group of strategies (\emph{Limited Label Availability}) is able to handle concept drift with a limited amount of to true labels whereas the second group of strategies (\emph{Unlimited Label Availability}) requires unlimited access to true labels. We start by explaining the four strategies within the first group.

%\paragraph{No retraining (No Retr.)}
The first benchmark is a non-adaptive model, \emph{No Retraining (No Retr.)}.
This strategy does not test for drifts and the ML model is only trained once with the initial training set.
The performance of this strategy constitutes a lower-bound benchmark. 

%\paragraph{Uninformed retraining (Uninf.)} 
The second benchmark is an \emph{Uninformed Retraining (Uninf.)} strategy which randomly draws retraining points out of all possible time stamps included in the respective data stream.
To ensure comparability, we set the number of retrainings of this strategy to be equal to the \emph{UDD} approach.
This also ensures that the uninformed retraining strategy receives access to the same number of true labels.
Otherwise, a strategy with access to more true labels will likely perform better due to larger training set sizes.
To get a reliable performance estimate for this strategy, we repeat this experiment five times and average the results. 

The third benchmark, \emph{Equal Distribution (Equal D.)}, is similar to the previous benchmark but the retraining points are equally distributed over the course of the data stream.

The \emph{Kolmogorov-Smirnov} test-based drift detector (\emph{KSWIN}) belongs to the category of input data-based drift detection and works by individually investigating each input feature for changes. 
We optimize its sensitivity parameter $\alpha$ with the same procedure as for \emph{UDD}. 
This detector is known to produce many false positive concept drift signals, due to multiple hypothesis testing (\cite{raab2020reactive}).
Again, we restrict the number of retrainings to be equal to the \emph{UDD} approach.
If this strategy detects more drifts, detected drifts are sorted by the order of their p-values and only the top drifts are considered for retraining.
For this strategy, we use the \texttt{scikit-multiflow} (\cite{skmultiflow}) implementation \emph{KSWIN} (Kolmogorov-Smirnov WINdowing) with the following parameters: $window\_size = 200$, $stat\_size=100$.

\subsubsection{Unlimited Label Availability}

The second group of strategies is not restricted with respect to the amount of allowed retrainings. 
Therefore, they are \textit{not} an appropriate benchmark in a context where true labels are scarce.
We still include these strategies since they serve as an upper-bound performance benchmark.
This allows us to estimate the performance loss when confronted with a situation where full label availability is infeasible.

%\paragraph{Kolmogorov-Smirnov test unlimited (KSWIN(unl.))} 
The \emph{Kolmogorov-Smirnov} test with \emph{unlimited} retrainings (\emph{KSWIN(unl.)}) benchmark is similar to the previous \emph{KSWIN} strategy but without restricting the number of retrainings. 
Therefore, all detected drifts trigger a retraining of the prediction model.

The last benchmark is the \emph{ADWIN} change detection algorithm applied to the prediction error rate.
%
%In case a drift is detected, the model is retrained with the recent data instances (one percent) and the training data.
%
This strategy already requires all true labels for the computation of the error rate and therefore for drift detection.
Note that all other strategies manage the drift detection without any true labels and then only require labels for retraining.
%
% Since this strategy requires all labels for computing the error rate in any case, we also do not limit the number of retrainings.
%
%Therefore, this strategy overall might use more true labels for adaptation of the model compared to the other strategies which naturally leads to a better prediction performance.
%
For this method, we use the \texttt{scikit-multiflow} (\cite{skmultiflow}) implementation with default parameter settings.

%\begin{itemize}
%    \item Explain used datasets
%    \item Explain difficulties when evaluating concept drift on real-world datasets since there is no ground truth available
%    \item Explain baselines
%    \item Show results
%    \item (Include detected drifts with uncertainty?)
%\end{itemize}

%
%
%
\subsection{Data Sets}
For evaluation, we consider two synthetic data sets (Friedman and Mixed) and ten real-world data sets. All data sets are widely used in concept drift research and are therefore suitable for evaluating \textit{UDD}. The \textbf{Friedman} regression data set (\cite{friedman1991multivariate}) consists of ten features that are each drawn from a uniform distribution from the interval $[0,1]$. %$\mathcal{U} \sim (0,1)$. 
The first five features are relevant for the prediction task, the remaining five are noise. 
%
%The number of instances is set to 20,000. The target variable is calculated as:
%\begin{equation}
%y = 10*\sin(\pi*x_1*x_2)+20(x_3-0.5)^2+10x_4+5x_3+e\sim \mathcal{N}(0,1)\,.
%\end{equation} 
%
The \textbf{Mixed} classification data set is inspired by (\cite{gama2004learning}) and contains six features where two features are Boolean and the other four features are drawn from a discrete distribution. 
Two of the features are noise which do not influence the classification function.
By modifying the distribution of some features, we can either induce real or virtual concept drifts (see Section \ref{sec:cd_definition}) in both the Friedman and the Mixed data set.

% and the number of instances is set to 20,000. $x_1$ and $x_2$ are Boolean features with equal probability for $True$ and $False$. The other features are drawn from the discrete distribution shown in Table \ref{table:distribution_mixed}.
% %
% \begin{table*}
% \caption{Distribution for features $x_3$, $x_4$, $x_5$, and $x_6$.}
% \label{table:distribution_mixed}
% \centering
% %\resizebox{\textwidth}{!}{
% \begin{tabular}{l c c c c c c c c c c} 
% \toprule
%     Value & 0.1  & 0.2 & 0.3 & 0.4 & 0.5 & 0.6 & 0.7 & 0.8 & 0.9 & 1  \\
%     Probability & 0.1 & 0.7 & 0.1 & 0.0125 & 0.0125 & 0.0125 & 0.0125 & 0.0125 & 0.0125 & 0.0125  \\
%     \bottomrule
% \end{tabular}
% %}
% \end{table*}
% %
% %\begin{align}
% %a &= [0, 0.1, 0.2, 0.3, 0.4, 0.5, 0.6, 0.7, 0.8, 0.9, 1] \\
% %p &= [0.1, 0.7, 0.1, 0.0125, 0.0125, 0.0125, 0.0125, 0.0125, 0.0125, 0.0125, 0.0125]
% %\end{align}
% %
% The class label is assigned according to the following rule:
% \begin{align}
% condition &= (x_4 < 0.5 +0.3*\sin(3\pi x_3)) \\
% y &= 
% \begin{cases}
% 1,\ if \ at \ least \ 2 \ of \ x_1, x_2 \ and \ condition\ are\ True \\
% 0,\ else
% \end{cases}
% \end{align}

Furthermore, ten real-world data sets---eight classification and two regression tasks---are used for the evaluation of the \emph{UDD} method.
%
%The characteristics of the real-world data sets regarding sample size, number of features, and targets can be found in the appendix in Table \ref{table:characteristics}.
%
The \textbf{Air Quality} data set (\cite{de2008field}) contains measurements from five metal oxide chemical sensors, a temperature, and a humidity sensor.
The learning task is to predict the benzene concentration, which is a proxy for air pollution.
%
%The real benzene concentration is measured with a specialized, expensive sensor.
%
Concept drift is present due to seasonal weather changes.
The \textbf{Bike Sharing} data set (\cite{fanaee2013event}) provides hourly rental data for a bike sharing system in Washington, D.C., with the objective to predict the hourly demand for bike rentals.
Concept drift is again assumed to be present due to seasonal weather changes.

All following classification data sets are taken from the USP Data Stream Repository (\cite{souza2020challenges}):
The various \textbf{Insects} data sets were gathered by controlled experiments on the use of optical sensors to classify six types of different flying insects.
Concept drift is artificially induced by changes in temperature.
The \textbf{Abrupt} data set contains five sudden changes in temperature, whereas in the Incremental (\textbf{Inc}) data set, temperature is slowly increased over time.
The Incremental Abrupt (\textbf{IncAbr}) data set has three cycles of incremental changes with additional abrupt drifts included as well.
In the Incremental Reoccurring (\textbf{IncReo}) data set, the temperature increases incrementally within several cycles.
The \textbf{KDDCUP99} data set %\cite{kdd99} was gathered at the MIT Lincoln Labs over a period of several weeks and 
contains TCP connection records from a local area network. 
%
%The full data set consists of five million connection records, however, we consider a 10\% subset, which is common practice \cite{souza2020challenges}. 
%Each record consists of 34 numerical and seven categorical features. 
%Applying one-hot encoding gives a total of 118 features
%
%Features comprise information such as connection duration, protocol type, and transmitted bytes.
%
The learning task is to recognize whether the connection is normal or relates to one of 22 different types of attacks.
The \textbf{Gas Sensor} data set contains records where one of six gases is diluted in synthetic dry air, and the objective is to identify the respective gas.
%\cite{vergara2012chemical} 
%was collected over 36 months at a gas delivery system.
%
%The sensor array is equipped with 16 chemical sensors measuring eight features each, resulting in a total of 128 features. 
%
%For each recording, one of six gases is diluted in synthetic dry air inside a sensing chamber, and the objective is to identify the respective gas.
%
Both sensor drift (due to aging) and concept drift (due to external alterations) are included in the data.
The \textbf{Electricity} data set 
%\cite{harries1999splice}
was gathered at the Australian New South Wales Electricity Market.
%
%Each record contains information about recent electricity consumption and market prices.
%
The learning task is to predict whether the market price will increase or decrease compared to the last 24 hours. %WHY DRIFT? SUITABILITY?
The \textbf{Rialto Bridge} Timelapse data set contains
%\cite{losing2016knn} 
images taken by a webcam close to the Rialto Bridge in Venice, Italy.
The objective is to correctly classify nearby buildings with concept drift occurring due to changing weather and lighting conditions.
\subsection{Performance Metrics}
Evaluating concept drift detection on real-world data sets is a challenging endeavor as most real-world data sets do not have specified drift points.
Specifically, for most real-world data, it is intractable to measure the accuracy of drift detection itself.
Therefore, we perform two different analyses regarding the behaviour of \emph{UDD} in this work. First, we apply \emph{UDD} on two synthetic data sets to specifically evaluate its drift detection capabilities. Second, we perform extensive experiments to investigate its performance on ten real-world data sets.

For synthetic data sets, the real drift points are known, which allows to compute metrics regarding the drift detection capabilities of a drift detector (\cite{bifet2013cd}).
In this work, we compute the Mean Time to Detection (MTD), the False Alarm Count (FAC), and the missed detection count (MDC).

In contrast, an appropriate evaluation for real-world data sets is more difficult.
However, one can assume that a drift is present when the prediction performance of a static model decreases over time.
Since the real drift points are unknown, we evaluate the different strategies based on their prediction performance, as it is common in the concept drift literature (\cite{elwell2011incremental}). 
For regression tasks, we apply the Root Mean Squared Error (RMSE), and we use the Matthews Correlation Coefficient (MCC) for classification tasks. MCC is a popular metric for classification settings as it can also handle data sets with class imbalance
(\cite{chicco2020advantages}).
%MCC is the metric of choice when it comes to classification \cite{chicco2020advantages}

%Symmetric Mean Absolute Percentage Error (SMAPE):
%Table \ref{table:baselines} shows the baseline for the two datasets. We compute the baseline as:
%\begin{equation} \label{e:baseline}
%SMAPE = \frac{100}{N} \sum_{i=1}^{N} \frac{2 * \left| \widehat{y}_i - y_i %\right|}{\left|\widehat{y}_i \right|+\left|y_i\right|}\,.
%\end{equation}

% Classification tasks are evaluated with the Area Under the Receiver Operating Characteristic Curve (AUROC) score.
% %
% The AUROC score takes both true positive and false positive rates into account and is therefore well-suited to deal with multi-class and imbalanced datasets.
% %
% An AUROC score of 0.5 is equivalent to the performance of a random guess.
% %

%
%. Thus, 0.5 serves as a baseline for the classification datasets.
% \begin{table}[h!]
% {\footnotesize
%  \caption{Baselines for regression datasets}
%  \label{table:baselines}
% \begin{center}
% \begin{tabular}{l|c} 

%  \textbf{Dataset} & \textbf{Baseline} \\
%  \hline
%  Air Quality & 61.04  \\ 
%  Bike Sharing & 85.06  \\
%  \hline
%  \end{tabular}
% \end{center}
% }
% \end{table}

\subsection{Analysis on Synthetic Data Sets}
To test the capabilities of \emph{UDD}, we analyze its behaviour when applied on two synthetic data sets (Friedman and Mixed).
Both data sets contain virtual as well as real concept drifts.
Virtual drifts refer to changes in the input data with no effect on the resulting label.
Hence, \emph{UDD} should \emph{not} raise an alarm for these drifts as a retraining of the ML model in this case is unnecessary.
Recall that this kind of analysis is only feasible on synthetic data sets, as we do not have any knowledge regarding the type of concept drift as well as its timing on real-world data sets.
On the synthetic data sets, we test \emph{UDD} and \emph{KSWIN(unl.)} as they both do not require true labels for drift detection.
The parameters of both approaches are optimized based on a validation set which includes one drift (see Section \ref{sec:experimental_setup}). 

\begin{figure}[t]
\centering
\includegraphics[width = \linewidth]{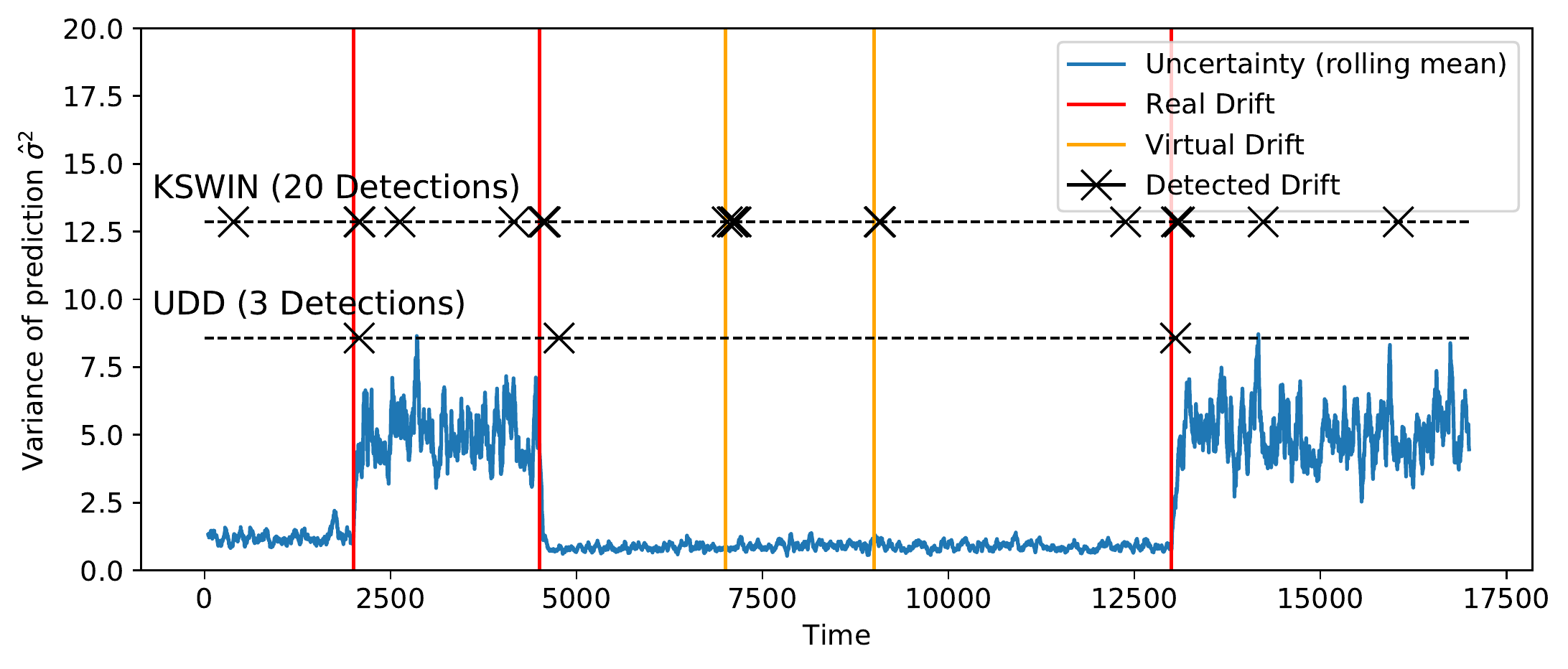}
\caption{Behaviour of \textit{UDD} and \textit{KSWIN} on synthetic Friedman data set.}
\label{fig:friedman}
\end{figure}

Figure \ref{fig:friedman} shows the trajectory of the predictive uncertainty over the course of the Friedman data set.
The uncertainty changes significantly each time a real concept drift occurs.
Accordingly, this is also detected by \emph{UDD}.
As expected, the two virtual drifts (marked by orange vertical lines in the figure) do not trigger a drift detection.
In contrast, the input data-based detection (\emph{KSWIN}) detects also these virtual drifts.
Furthermore, note the overall large number (20) of detected drifts by \emph{KSWIN} despite a parameter optimization.
This illustrates \emph{KSWIN}'s problem of high reactivity leading to several false-positive drift detections.
%
%Figure \ref{fig:mixed} in the appendix depicts this analysis for the Mixed data set.

\begin{table}[b]
\caption{Evaluation on synthetic data sets.}
\label{table:synthetic}
\centering
\resizebox{0.9\linewidth}{!}{
\begin{tabular}{l c c c c}
\textbf{Data Set} & \textbf{Drift Detector} & \textbf{MTD} & \textbf{FAC} & \textbf{MDC} \\
\toprule
\multirow{2}{*}{Friedman}   & \textit{UDD}   & 132.7 & 0   & 0   \\
                            & \textit{KSWIN} & 65.7  & 17  & 0   \\
%\hline
\multirow{2}{*}{Mixed}      & \textit{UDD}   & 247.3 & 0   & 0   \\
                            & \textit{KSWIN} & 50.3  & 11  & 0   \\
\bottomrule
\end{tabular}
}
\end{table}

Since the real drift points for the synthetic data sets are known, we compute the mean time to detection (MTD), the false alarm count (FAC), and the missed detection count (MDC) for both strategies in Table \ref{table:synthetic}.
\textit{UDD} correctly identifies all real concept drifts in both data sets.
Furthermore, no false alarms are raised.
However, \emph{KSWIN} achieves lower MTD values compared to \emph{UDD} in both data sets, which means that \emph{KSWIN} recognizes concept drifts faster.
This can likely be explained by the high sensitivity of \emph{KSWIN} regarding changes.
However, this sensitivity also leads to large numbers of false alarms (17 and 11, respectively), as depicted in Table \ref{table:synthetic}.
Such a behaviour is especially detrimental in scenarios where the acquisition of true labels is expensive.
Each time a false alarm is raised, new true labels must be acquired at a high cost---even though a retraining is not required since no real concept drift has occurred.

\subsection{Experimental Results}
Both \emph{UDD} and \emph{KSWIN} require as input a suitable value for $\alpha$, which determines their sensitivity regarding concept drift detection.
Since the data sets included in this experiment are fundamentally different from each other (e.g., different number of class labels), individual values of $\alpha$ are required for each data set.
As described in Section \ref{sec:experimental_setup}, we determine the respective value for both strategies by performing a test on a validation data set. 
%The parameter values for \emph{UDD} for each data set are depicted in Table \ref{table:adwin-alpha} in the appendix.
%
\begin{table*}[h]
\caption{RMSE (the lower the better) on \textit{regression} benchmark data sets. Number of retrainings in brackets (the lower the less computationally expensive). \textit{No Retraining} depicts the lower-bound benchmark, while \textit{KSWIN(unl.)} and \textit{ADWIN} represent the upper-bound performance benchmark.}
\label{table:results_regression}
\centering
\resizebox{0.85\linewidth}{!}{
\begin{tabular}{l c c c c c c c} 
    & \multicolumn{5}{c}{\textbf{Limited Label Avail.}} & \multicolumn{2}{c}{\textbf{Unlimited Label Avail.}} \\
    \textbf{Data Set} & \textbf{No Retr.} & \textbf{Uninf.} & \textbf{Equal D.} & \textbf{KSWIN} & \textbf{UDD} & \emph{\textbf{KSWIN(unl.)}} & \emph{\textbf{ADWIN}} \\
    \toprule
    Air Quality & 1.170 (0)   & 1.383 (14)  & 1.231 (14) &   1.285 (14) & \textbf{1.151} (14)  & 1.304 (19) & 1.387 (12)   \\
    Bike Sharing & 171.47 (0) & 170.00 (5)  & 144.94 (5)    &  143.88 (5)    & \textbf{129.93} (5) & 120.69 (27) & 127.07 (8)  \\
%    \textit{Friedman (syn)}    & 3.950 (0) & 3.229 (3)  & 3.476 (3) & 3.630 (3) & \textbf{2.111} (3) & 1.674 (20) & 2.080 (3) \\
    \bottomrule
\end{tabular}
}
\end{table*}
\begin{table*}[h]
\caption{MCC (the higher the better) on \textit{classification} benchmark data sets. Number of retrainings in brackets (the lower the less computationally expensive). \textit{No Retraining} depicts the lower-bound benchmark, while \textit{KSWIN(unl.)} and \textit{ADWIN} represent the upper-bound performance benchmark.}
\label{table:results_classification}
\centering
\resizebox{0.85\linewidth}{!}{
\begin{tabular}{l c c c c c c c} 
    & \multicolumn{5}{c}{\textbf{Limited Label Avail.}} & \multicolumn{2}{c}{\textbf{Unlimited Label Avail.}} \\
    \textbf{Data Set} & \textbf{No Retr.} & \textbf{Uninf.} & \textbf{Equal D.} & \textbf{KSWIN} & \textbf{UDD} & \emph{\textbf{KSWIN(unl.)}} & \emph{\textbf{ADWIN}} \\
    \toprule
    Insects Abrupt & 0.452 (0)   & 0.468 (9)  & 0.475 (9) &   0.456 (9) & \textbf{0.516} (9)  & 0.521 (192) & 0.497 (9)   \\
    Insects Inc & 0.052 (0) & 0.210 (4)  & 0.211 (4)    &  0.191 (4)    & \textbf{0.242} (4) & 0.238 (27) & 0.251 (3)  \\
    Insects IncAbr & 0.292 (0) & 0.463 (22) & 0.483 (22) & 0.464 (22)   & \textbf{0.522} (22) & 0.488 (107) & 0.516 (23) \\
    Insects IncReo & 0.114 (0) & 0.190 (10) & 0.197 (10) & 0.126 (10)   & \textbf{0.208} (10) & 0.218 (149) & 0.239 (13)  \\
    KDDCUP99    & 0.663 (0) & 0.830 (20)  & 0.873 (20) & 0.772 (20)         & \textbf{0.964} (20) & 0.986 (345) & 0.984 (61)  \\
    Gas Sensor     & 0.255 (0) & 0.472 (39) & 0.469 (39) & 0.325 (39) & \textbf{0.484} (39) & 0.454 (149) & 0.480 (49)  \\
    Electricity & 0.139 (0) & 0.372 (13)  & 0.362 (13) & 0.254 (13) & \textbf{0.436} (13) & 0.511 (269) & 0.471 (45) \\
    Rialto Bridge   & 0.534 (0)  & 0.558 (14)  & 0.561 (14) & \textbf{0.583} (14)  & \textbf{0.583} (14) & 0.586 (17) & 0.600 (116)  \\
%    \textit{Mixed (syn)}    & 0.875 (0) & 0.924 (3)  & 0.897 (3) & 0.947 (3) & \textbf{0.952} (3) & 0.952 (14) & 0.955 (3) \\
    \bottomrule
\end{tabular}
}
\end{table*}

A summary of the experimental results on all data sets is provided in Table \ref{table:results_regression} for regression data sets (RMSE) and in Table \ref{table:results_classification} for classification (MCC). 
%
%Furthermore, we provide an additional view on the results by depicting the SMAPE metric in Table \ref{table:results_regression_smape} and the F1-score in Table \ref{table:results_classification_f1} in the appendix.
%
For the evaluation, we primarily focus on the first five columns of the table which as a group can be characterized by only requiring a limited amount of true labels. 
This is also illustrated by the values in parentheses which describe how often the corresponding ML models are retrained.  
As explained in Section \ref{sec:experimental_setup}, \emph{KSWIN(unl.)} and \emph{ADWIN} serve as an upper-bound benchmark due to their requirement of full label availability.

The best strategy with limited label availability per data set is marked in bold.
For both regression data sets, \emph{UDD} outperforms the other four strategies.
Regarding the classification tasks, \emph{UDD} achieves the best prediction performance on seven out of eight data sets and always outperforms the strategies \textit{No Retraining},  \textit{Uninformed} and \textit{Equal Distribution}.
Solely for the Rialto data set, the strategy based on \emph{KSWIN} performs equally well, which might be explained with rather significant changes in individual input features that can be detected well with \emph{KSWIN}.
As expected, the \textit{No Retraining} strategy usually performs worst.
This finding clearly illustrates the presence of concept drift in all of the selected real-world data sets even though the exact drift points are not measurable.
Interestingly, the \textit{Uninformed} already achieves good prediction performance and sometimes even outperforms the \textit{KSWIN} strategy, especially for the regression tasks.
By design, the number of retrainings is equal for all four strategies---\textit{Uninformed}, \emph{Equal Distribution}, \textit{KSWIN}, and \emph{UDD}.

The right two columns in both Table \ref{table:results_regression} and Table \ref{table:results_classification} show the prediction performance of the \textit{KSWIN(unl.)} and \textit{ADWIN} strategy.
As expected, these strategies usually outperform all other strategies but also require significantly more true labels for retraining. For the KDDCUP99 data set, the difference in amounts of retrainings for \emph{UDD} compared to \emph{KSWIN(unl.)} is most striking: While \emph{UDD} requires 16 retraining, \emph{KSWIN(unl.)} performs 345 retrainings in total. Yet, the difference in predictive performance is rather small.
Also, recall that the \textit{ADWIN} strategy requires \textit{all} true labels for drift detection itself.
For the Insects Abrupt, Insects IncAbr, and the Gas Sensor data set, the \emph{UDD} strategy performs even better than \textit{ADWIN}.
%
%In that case, \emph{UDD} might be better at detection because the uncertainty values are not restricted to a binary distribution as it is the case for \emph{ADWIN} with classification errors.

% \begin{table}
% \caption{Correlations between prediction error and uncertainty.}  \label{table:corr}
% \centering
% \begin{tabular}{l c} 
%     \textbf{Dataset} & \textbf{Correlation} \\
%     \toprule
%     Air Quality & 0.4248  \\ 
%     Bike Sharing & 0.3670  \\
%     \hline
%     Insects Abrupt & 0.5335  \\
%     Insects Incremental & 0.3038  \\
%     Insects Inc-Abr & 0.6268  \\
%     Insects Inc-Reoc & 0.5674 \\
%     KDDCUP99 & 0.8340  \\
%     Gas Sensor Array & 0.7399 \\
%     Electricity & 0.4126  \\
%     Rialto Bridge & 0.4343  \\
%     \bottomrule
% \end{tabular}
% \end{table}

% We also investigate the relationship between prediction error and the uncertainty metric of \emph{UDD} by computing Spearman's rank correlation coefficient.
% %
% For regression tasks, we use the Mean Absolute Error (MAE) as prediction error and the negative log-loss error for classification tasks.
% %
% Table \ref{table:corr} shows the correlation coefficients per dataset, which all indicate a positive correlation and thus confirm the initial assumption that model uncertainty can be used as a proxy for the error measure.
% Furthermore, the lowest correlation is found for the Insects Incremental dataset, which is one of the two cases ADWIN Uncertainty was outperformed by KSWIN. This indicates

\begin{figure*}[h]
\centering
    \subfloat[Regression]{\label{fig:a}\includegraphics[width=0.4\linewidth]{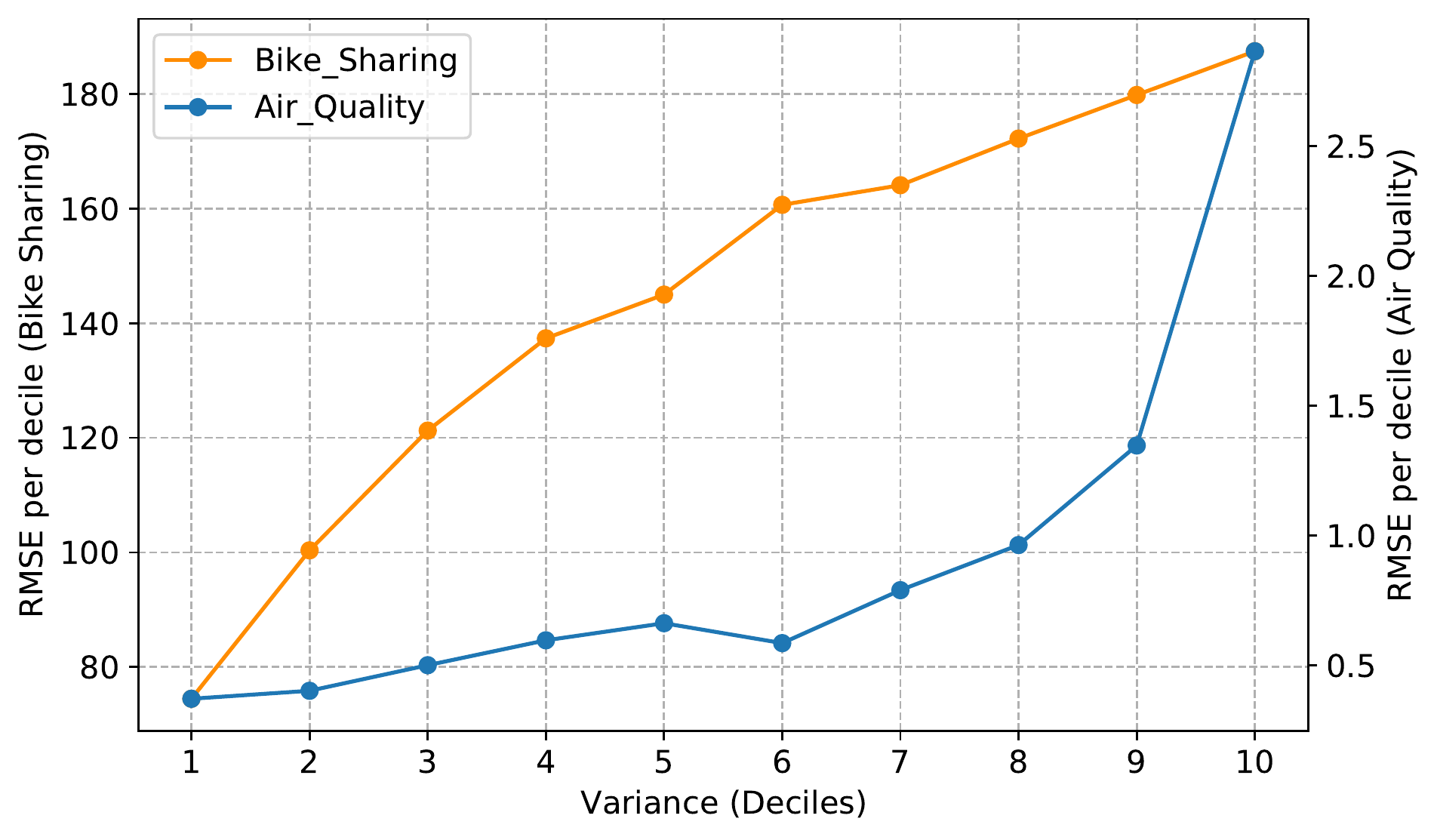}}
    % \\
    \subfloat[Classification]{\label{fig:b}\includegraphics[width=0.37\linewidth]{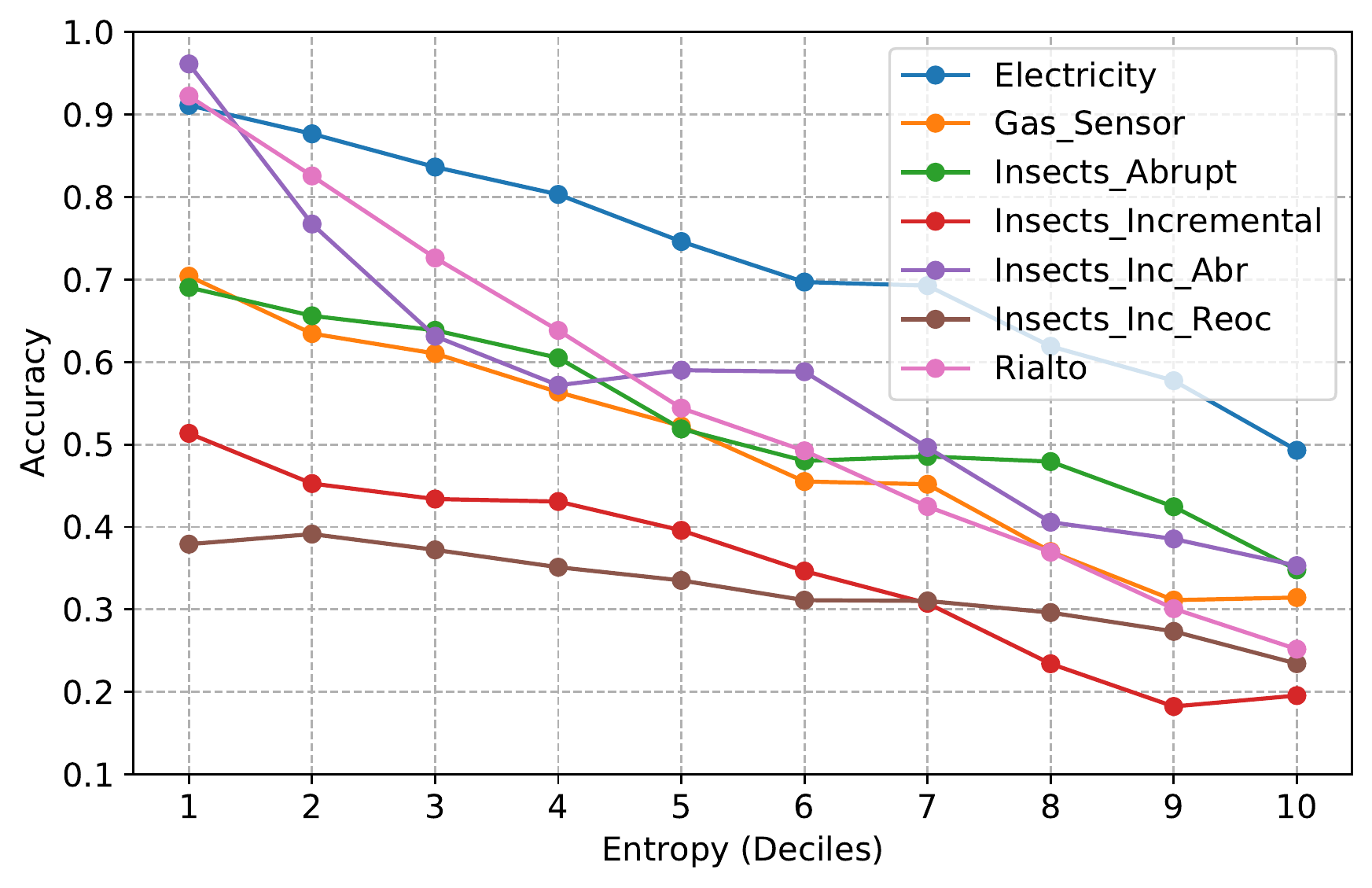}}
\caption{Relationship between deciles of uncertainty and prediction performance.}
\label{fig:error-vs-unc}
\end{figure*}

We also investigate the average prediction performance for \emph{UDD} based on the level of uncertainty in Figure \ref{fig:error-vs-unc}.
Per data set, we sort instances in deciles, from instances with lowest uncertainty (decile 1) up to instances with highest uncertainty (decile 10) based on entropy $H$ or variance $\widehat{\sigma}^{2}$, respectively.
Subsequently, we compute the average prediction performance per decile.
As expected, the RMSE for regression data sets increases with rising uncertainty, as shown in the left plot (a).
The right plot (b) shows the classification data sets---decile 1 shows the highest mean accuracy and decile 10 the lowest.\footnote{ KDDCUP99 data set is not included in Figure \ref{fig:error-vs-unc} 
%and \ref{fig:error-vs-conf} 
because deciles cannot be computed due to the skewed entropy distribution.} Thus, Figure \ref{fig:error-vs-unc} confirms our assumption that uncertainty represents a proxy for the error metric.

\begin{comment}

\begin{figure}
\centering
\includegraphics[width=0.75\linewidth]{figures/Accuracy_Confidence.pdf}
\caption{Relationship between deciles of confidence and accuracy.}
\label{fig:error-vs-conf}
\end{figure}

\end{comment}

%For classification tasks, we additionally analyze the relationship between confidence and accuracy (similar to \cite{lakshminarayanan2017simple}). We define confidence $c := \max_{k} \widehat{p}(y = k \vert x)$ as the highest predicted probability for one class of a specific data instance. All instances of a data set are sorted into confidence deciles, where decile 10 contains all instances with highest confidence. Figure \ref{fig:error-vs-conf} depicts the relationship. As expected, the mean accuracy score increases with larger confidence.
%
\section{Conclusion}
In this work, we have introduced the \emph{Uncertainty Drift Detection (UDD)} algorithm for concept drift detection.
As stated in the research question, this algorithm is also suitable for situations with limited availability of true labels since it does not depend on true labels for detection of concept drift.
Only in case of a detected drift, it requires access to a limited set of true labels for retraining of the prediction model.
Therefore, this algorithm is especially suitable for drift handling in deployed ML settings within real-world environments where the acquisition of true labels is expensive (e.g., quality control).
Standard drift detection algorithms such as DDM and ADWIN are not applicable in such settings because they require access to the entire set of true labels.
Our approach is based on the uncertainties derived from a deep neural network in combination with Monte Carlo Dropout.
Drifts are detected by applying the ADWIN change detector on the stream of uncertainty values over time.
In contrast to most existing drift detection algorithms, our approach is able to detect drift in both regression and classification settings.
We have performed an extensive evaluation on two synthetic as well as ten real-world concept drift data sets to demonstrate the effectiveness of \emph{UDD} for concept drift handling in comparison to other state-of-the-art strategies.

However, more evaluation of \emph{UDD} in various use cases with different data sets is required to prove its overall effectiveness. 
Additionally, \emph{UDD} can only be applied successfully for use cases where concept drift can be observed in the input data---as opposed to drift in the labels alone. 
In future work, we aim to improve the \emph{UDD} method by including active learning methods.
Including only those instances with high uncertainty in the retraining set rather than all recent instances could further improve the prediction performance. 
Furthermore, we also want to analyze which type and magnitude of concept drift can best be handled by applying \emph{UDD}.

\printbibliography

\end{document}

%% file: hicss-packages.tex
\usepackage[letterpaper]{geometry}
\usepackage{hicss}
\usepackage{times}
\usepackage[none]{hyphenat}
\usepackage{url}
\usepackage{latexsym}
\usepackage{indentfirst}
\usepackage{graphicx}
\graphicspath{{images/}}
\usepackage[
    style=apa,
  ]{biblatex}